\documentclass[12pt]{article}

\usepackage[bitstream-charter]{mathdesign}
\usepackage[mathcal]{eucal}

\usepackage[english]{babel}

\usepackage{pifont}
\usepackage{xcolor}
\usepackage{graphicx}
\usepackage{amsmath}
\usepackage{multirow}
\usepackage{blindtext}
\usepackage{wrapfig,booktabs}
\usepackage{mathtools}
\usepackage{comment}
\usepackage{lipsum}
\usepackage{enumitem}
\usepackage[colorlinks=true,citecolor=blue]{hyperref}
\usepackage{adjustbox}
\usepackage{subcaption}
\usepackage{algpseudocode}
\usepackage{algorithm}
\usepackage{a4wide}
\usepackage{fancyhdr}
\usepackage{authblk}
\usepackage{emptypage}

\definecolor{mycolor}{rgb}{0.122, 0.435, 0.698}

\newcommand{\nome}{Pixle}

\definecolor{bostonuniversityred}{rgb}{0.8, 0.0, 0.0}

\fancypagestyle{plain}{%
    \fancyhf{}
  \rhead{Pre-print submitted for review to an \textbf{IEEE Conference}}
}

\begin{document}

\title{\nome{}: a fast and effective black-box attack \\ based on rearranging pixels}

\author[1,*]{Jary Ponponi}
\author[1]{Simone Scardapane}
\author[1]{Aurelio Uncini}

\affil[1]{\small Sapienza University of Rome, Via Eudossiana 18, 00184, Rome, Italy}
\affil[*]{Corresponding author email: jary.pomponi@uniroma1.it}

\maketitle

\begin{abstract}
	Recent research has found that neural networks are vulnerable to several types of adversarial attacks, where the input samples are modified in such a way that the model produces a wrong prediction that misclassifies the adversarial sample. In this paper we focus on black-box adversarial attacks, that can be performed without knowing the inner structure of the attacked model, nor the training procedure, and we propose a novel attack that is capable of correctly attacking a high percentage of samples by rearranging a small number of pixels within the attacked image. We demonstrate that our attack works on a large number of datasets and models, that it requires a small number of iterations, and that the distance between the original sample and the adversarial one is negligible to the human eye. 
\end{abstract}

	\section{Introduction}
    
    Neural Networks (NN) have achieved state-of-the-art performance in image classification and many other tasks. However, despite their extensive usage, NNs are highly susceptible to deception performed using adversarial images, which are samples containing small noise, used to fool the NN and force a misclassification. Models that operate in a real world scenario can also be attacked, by physically modifying the objects that have to be classified \cite{athalye2018synthesizing, eykholt2018robust}. This problem is not circumscribed to NNs that operate in the image domain, but also to ones that work over speech, text, or more general domains.  
    
    The majority of the existing methods to generate an adversarial image do it by constraining the difference between the original image and the adversarial one, in such a way that the difference between the two is small, and thus hardly visible to the naked eye. The difference between these images is usually calculated using a norm $L_p$. The value of $p$ influences the method used by the algorithm, and we can have: $L_0$) the difference is pixel wise, by counting the number of pixels that have been modified by the method, $L_1$) which measures the absolute distance between the images, $L_2$) that measures the distance as Euclidean distance and, $L_\infty$) which is the largest perturbed pixel in the image.    
    
    Another important aspect when developing an attack is the number of iterations required to correctly create an adversarial image. An iteration is an interrogation of the model: the input is passed though it and the output is collected. It is a crucial aspect, because many real life models have a limit on the number of queries that can be performed, and thus it is important to keep this aspect contained.  
    
    \begin{figure}[t]
    \centering
    \includegraphics[width=0.5\linewidth]{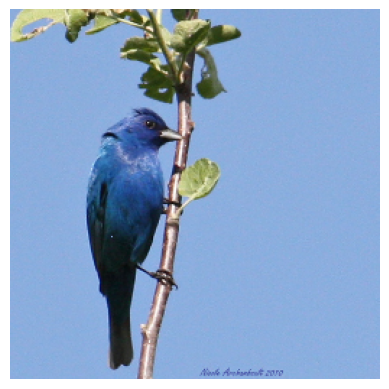}
    \caption{Successful adversarial image from ImageNet using the proposed \nome{} attack. The original class of the image is 14 (Passerina Cyanea), while the misclassified class is 883 (Vase). The $L_0$ distance between the original image and the adversarial one is just $1$ pixel in this case.}
    \label{fig:introduction}
    \end{figure}
    
    Moreover, adversarial methods can be grouped into two categories: white-box attacks and black-box attacks \cite{bhambri2019survey, app9050909}. In the first case, which is the most studied one, the methods have full knowledge about the model and its training procedure. In this case, we can rely on any information within the model itself, such as gradients for a given sample or the weights of the model itself. The methods in this category clarify the risks to which the NNs may be exposed, but usually they do not reflect a real case scenario, in which a malicious user wants to attack a model, but has a limited number of queries and no access to the model itself, with the exception of its output. On the other hand, in the black-box case we can access only the input and the output of the model, and thus it is closer to a true case scenario. 

    In this paper we propose a simple yet effective $L_0$ norm black-box attack. Our proposal, called \nome{}, is capable of efficiently attacking a model by measuring only, given an input sample, the confidence of the prediction, as a probability. \nome{} is based on random search, and it creates an effective adversarial image by rearranging a small set of pixels in the image, showing also that an image usually contains all the pixels that are necessary to misclassify it, using a contained number of iterations, as shown in our experimental results.
    
	\section{Related Works}
    
    The problem of security in machine learning has always been crucial in order to develop more robust models \cite{barreno2010security, barreno2006can}. The first machine learning model to be attacked were Support Vector Machines \cite{biggio2012poisoning}, then, in \cite{szegedy2014intriguing} and \cite{goodfellow2014explaining}, the authors discovered that NNs are prone to such attacks as well, which can be easily accomplished using several gradient based algorithms for obtaining gradient information of the attacked image, and thus fool the model. After these discoveries, the security of NNs has become a critical topic for real world deployments. 
    
    Many methods to fool a NN have been proposed in recent years. As said, white-box attacks are the most studied. As attacks based on $L_0$ norm we have SparseFool \cite{modas2019sparsefool}, which exploits the curvature of decision boundaries, or JSMA \cite{papernot2016limitations}, that finds the pixels to attack through saliency maps. Based on others norms, we have the Fast Gradient Sign Method \cite{goodfellow2014explaining} and its evolution, called Projected Gradient Descent \cite{madry2018towards}, and Jitter \cite{schwinn2021exploring}, for $L_\infty$ norm; both methods attack the image by changing all the pixels based on their importance, calculated using the gradient associated to the image. For $L_1$ norm we have attacks that place an upper bound over the absolute sum of the perturbed values, and the most notable examples are DeepFool \cite{moosavi2016deepfool}, EAD \cite{sharma2017attacking}, and FAB \cite{croce2020minimally}. 
    
  Black-box attacks are usually based on $L_2$ or $L_\infty$ norms. In \cite{8014906} the authors proposed a simple policy that produces adversarial images by perturbing random segments of an image to produce the adversarial counterpart. In \cite{8014906}, the authors proposed the idea to identify low-frequency perturbations to improve query efficiency when attacking a model. Another group of methods is composed by approaches that estimate the gradient of the model without accessing it directly \cite{ilyas2018black, alzantot2019genattack, bhagoji2017exploring}, emulating white-box attacks. Less studied black-box attacks are the ones based on the $L_0$ norm. The first method developed in this field is called One-Pixel attack \cite{su2019one}, that uses a Differential Evolution (DE) search algorithm \cite{storn1997differential} to find the best pixels to replace and the values that must overwrite them. It works well on small images, but fails when it is not the case; also, it requires thousands of iterations to find a suitable set of pixels to replace. A similar approach, but based on larger patches, is called PatchAttack \cite{yang2020patchattack}, that uses reinforcement learning to optimally place pre-computed patches over the image to attack. The drawbacks are that the patches are easily detected, being, often, very large and visible.  Another approach, called ScratchThat \cite{jere2019scratch} is also based on DE, and literally `scratches' the images by adding lines of different colors over them to create perturbed images. 
    
    As said, studying how to fool NNs is important to understand also how to protect them from such attacks, and many approaches have been proposed over the years \cite{yuan2019adversarial}. One way to achieve this goal is to add adversarial images to the training data, such that the robustness against adversarial images can be improved against specific attacks \cite{huang2015learning, madry2018towards}. Others training approaches build more robust models by distillation \cite{papernot2016distillation}, ensemble of multiple models or models that incorporate some degree of uncertainty within them \cite{gawlikowski2021survey}. In addition,  some image processing methods are also proved to be effective in detecting adversarial images. For instance, in \cite{liang2018detecting} the authors studied how the detection of adversarial images can be carried out using noise reduction methods, by comparing the classification after and before applying those techniques. Similarly, in \cite{liang2018detecting} the authors showed that squeezing colors and applying spatial smoothing have high success rate on detecting adversarial images. 
    

	\section{Methodology}
    
    \begin{figure}[t]
        \centering
        \includegraphics[width=0.9\linewidth]{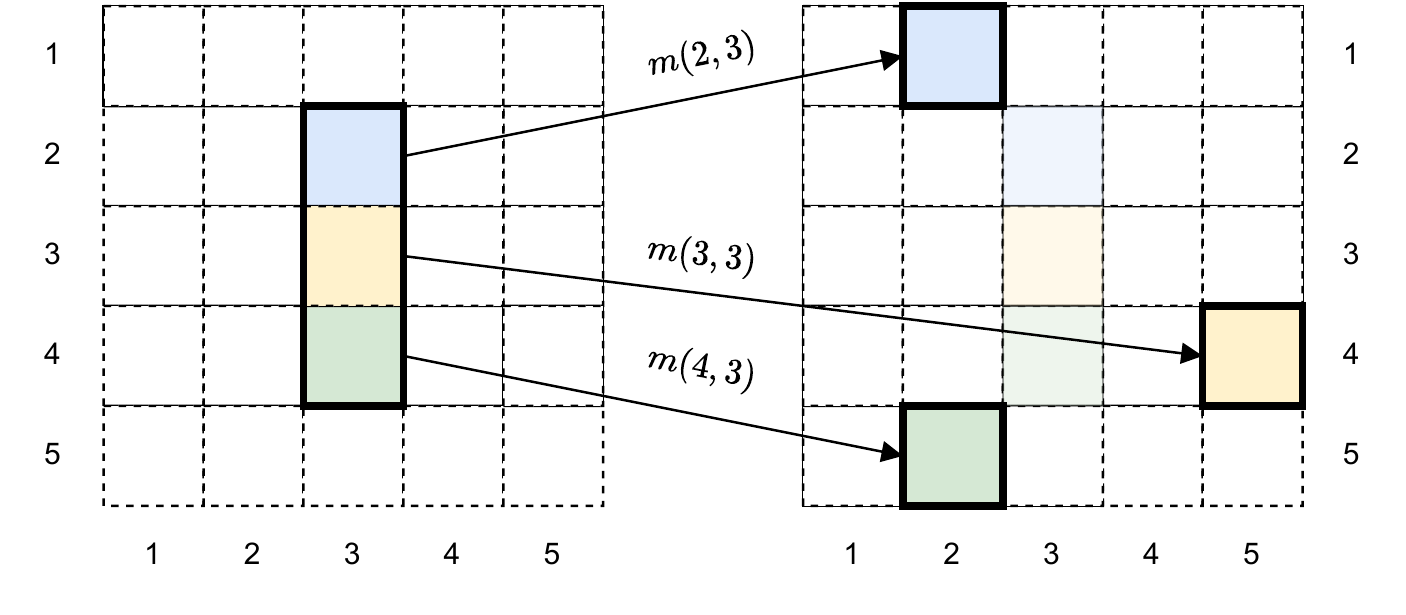}
        \caption{Visualization of the mapping procedure used by \nome{}. As the image shows, pixels originating from a small patch of the source image (left) are mapped into the destination image (right), to build the adversarial sample. The pixels in the patch are the only ones that are mapped into new positions, while the others are unchanged. Better viewed in color.}
        \label{fig:mappin}
    \end{figure}
    
    
	\subsection{Problem Description}
	Generating adversarial images that fool a neural network can be formalized as an optimization problem with constraints. 
    
	Let $f: x \in [0, 1] \in \mathbb{R}^{(3, w, h)} \rightarrow \mathbb{R}^c$ be a classification function that takes as input a sample of size $(3, w, h)$, were $3$ are the number of channels while $w$ and $h$ are, respectively, the width and the height of the image, and produces a vector containing $c$ probabilities (such that $\sum_i c_i = 1$), one associated to each class; a sample $x$ can be classified as $\arg \max_i f_i(x)$. In our case, the function $f$ is a NN. 
	
	Given an image $x$, and the associated class $y$, that is classified correctly by the function $f$, the goal of an attack is to change the prediction of the function, by producing an adversarial image $\overline{x} \in \mathbb{R}^{(3, w, h)}$ such that $f(\overline{x}) \ne y$. The image $\overline{x}$ must be similar to $x$, in such a way that the $L_p$ norm of the difference is bounded by a constant $\epsilon$: 
	
	\begin{equation}
	    \arg \max_i f_i(\overline{x}) \ne y \ \ \ \ \lvert \lvert x - \overline{x} \rvert \rvert_p \le \epsilon 
	\end{equation}
	
	\noindent with $\epsilon \in \mathbb{N}_+$ in our black-box case. An attack can be untargeted, as described above, or targeted, by forcing the miss-classification of the adversarial image to a specific adversarial class $\overline{y} \ne y$: $f(\overline{x}) = \overline{y}$. The task of finding the perturbed image $\overline{x}$, associated to $x$, can be viewed as a minimization problem
	
	\begin{equation}
	    \min_{\overline{x}} L(f(\overline{x}), y) \ \ \ \ \lvert \lvert x - \overline{x} \rvert \rvert_p \le \epsilon \,,
	\end{equation}
	
	\noindent for a proper loss function $L$. Here, we use $L(f(\overline{x}), y) = f_y(\overline{x})$ for untargeted attacks and $L(f(\overline{x}), \overline{y}) = 1 - f_{\overline{y}}(\overline{x})$ for targeted ones; in the first case we want to minimize the score associated to the correct class, while in the second one we want to maximize the score associated to the adversarial target $\overline{y}$. 
	
	\subsection{\nome}
    
	\nome{} is a black-box attack based on random search. Despite its simplicity, random search performs well in many situations and does not depend on gradient information associated to the objective function $L(\cdot, \cdot)$.
    
    Given an image $x \in \mathbb{R}^{(3, w, h)} $, the main idea of the algorithm is to sample a random portion $p$ of adjacent pixels from it, that we call \textit{patch}, and to rearrange some pixels inside it into other positions, calculated, for each pixel, using a predefined function $m$, that we call the \textit{map} function. Given an image $x$, a generic patch is defined by a $4$-tuple $p = (w_p, o_y, w_p, h_p)$, where $0 < o_x \le w$ and $0 < o_y \le h$ are the coordinates on the image used as origin point of the patch, while $w_p$ and $h_p$ are, respectively, the weight and the height of the patch. We indicate as $P$ the list of index tuples $(i, j)$, that are defined by the coordinates of the patch, and thus are contained in the rectangle having vertices as $[(o_x, o_y), (o_x + w_p, o_y), (o_x, o_y + h_p), (o_x + w_p, o_y + h_p)]$. We formalize this list as $P = \left[(o_x + i, o_y + j)\right]_{\forall i \in \{0,\dots,w_p\},j \in \{0,\dots,h_p\}}$, which has cardinality $\lvert P \rvert = w_p * h_p$, assuming that the patch does not exceeds the boundary of the image, otherwise it is moved in such a way that all the indexes are inside the image. The mapping function $m$ is defined as:
    
	\begin{equation}
    m\colon (i, j) \in P  \mapsto \mathbb{N}^2_+ \in [1, h] \times [1, w] \,,
	\end{equation}
        
    \noindent that returns, for each pixel in the patch, the position where the pixel must be moved in the original image. 
    Given this function, we can define each pixel of the adversarial image as:
    
	\[ 
	 \overline{x}_{i, j} = \begin{cases} 
      m(i, j) & (i, j) \in P \\
      x_{i, j} & \text{otherwise}
    \end{cases}
    \]

    The exposed procedure changes only the destination pixels, that are overwritten by the source ones. This approach is useful if we want to search for an adversarial image that reduces the distance from the original one. If we don't care about this aspect, and we want to speed up the convergence, another possible approach is to swap the position of the source and destination pixels at the same time, using the mapping function as before. In this way, no redundant information is injected into the adversarial image. The overall procedure, sampling a patch and calculating the mapping function codomain, remains the same. Fig. \ref{fig:mappin} visually shows how the pixels are mapped from the source image to the destination one.
    
	\begin{algorithm}[t]
    \caption{\nome{}: Restart-Iterative algorithm}
    \label{alg:restart}
    \begin{algorithmic}
    \Require input image $x$ with its associated label $y$. Maximum and minimum dimension for source patch. The number of restarts $R$ and the iterations to perform for each restart step $T$. The mapping function $m$.
    \Ensure $y = x^n$
    \State $\overline{x} \gets x$
    \State $l \gets f_y(x)$
    \For{$r = 0$ to $R$}
        \State $x^r \gets \overline{x}$
        \For{$t = 0$ to $T$}
            \State Sample $p = (o_x, o_y, w_p, h_p)$.
            \State Calculate the set $P$
            \State $x^{t} \gets \overline{x}$
            \For{$\forall (i, j) \in P$}
                \State $(z, k) \gets m(i, j)$
                \State $x^{t}_{z, k} \gets x_{i, j}$
            \EndFor 
        
        \If{$f_y(x^{t}) < l$}
            \State $l \gets f_y(x^{t})$
            \State $x^r \gets x^{t}$
        \EndIf
        \EndFor
        
        \State $\overline{x} \gets x^r$
        
        \EndFor
    
    \Return $\overline{x}$
    
    \end{algorithmic}
    \end{algorithm}
    
\subsection{Implementing the mapping function}
    
    The function $m$ can be implemented in different ways. Here, since we focus on the speed of the attack and the number of times that the models is interrogated, we propose and test the following implementations:
    
    \begin{itemize}
        \item Random: the function $m$ returns, for each input pixel, a random coordinate point $(i \sim \mathbb{U}[1, h], j \sim \mathbb{U}[1, w])$, different from the origin position.   
        \item Similarity: given the image $x$ and the pixel $i$ extracted using an index tuple from $P$, it returns the position of the most similar pixel in the image that is not equal to $i$. 
        \item Distance: it works like the Similarity approach, but in this case the position of the most different pixel is selected, avoiding the ones that have zero distance.
        \item Similarity Distribution: like the similarity approach, this is based on the most similar pixels, but, instead of selecting the position in a deterministic way, a distribution of positions is calculated using all the distances between the source pixel and all the others; in this way, the most similar pixels have a higher probability of being selected. Once the distribution has been calculated, the mapped position is sampled from it. 
        \item Distance Distribution: it is the distance based version of the Similarity Distribution exposed above, where most distant pixels are more likely to be selected.
    \end{itemize}
    
    
	
    
	\subsection{Search Algorithms}
    
	\begin{algorithm}[t]
    \caption{\nome{}: Iterative algorithm}
    \label{alg:iterative}
    \begin{algorithmic}
    \Require input image $x$ with its associated label $y$. Maximum and minimum dimension for source patch. The number of iterations to perform for each restart step $T$. The mapping function $m$.
    \Ensure $y = x^n$
    \State $\overline{x} \gets x$
    \State $l \gets f_y(x)$
    \State $x \gets \overline{x}$
    \For{$t = 0$ to $T$}
        \State Sample $p = (o_x, o_y, w_p, h_p)$.
        \State Calculate the set $P$
        \State $x^{t} \gets \overline{x}$
        \For{$\forall (i, j) \in P$}
            \State $(z, k) \gets m(i, j)$
            \State $x^{t}_{z, k} \gets x_{i, j}$
        \EndFor 
            
    \If{$f_y(x^{t}) < l$}
        \State $l \gets f_y(x^{t})$
        \State $\overline{x} \gets x^{i}$
    \EndIf
    \EndFor
    
    \Return $\overline{x}$
    
    \end{algorithmic}
    \end{algorithm}
    We propose two different procedures to find the adversarial image. In the first one, we have two loops, one nested inside the other, and we select only the attack that leads to the highest decrease of the loss; this attack is called Restart-Iterative. In the second one, each pixel replacement is used to create the final adversarial image, and it is called Iterative.
    
    
	\textbf{Restart-Iterative}: the algorithm is composed of a fixed number of restarts $R \ge 1$, and within each one a maximum number of iterations $M$ are performed. Before starting the main loop, the adversarial image is a copy of the original one, so that all the pixels from the original image are preserved during the process. At every iteration, we sample a source patch $p$, and then, using the mapping function $m$, the pixels in the set $P$ are overwritten. Only the mapping that decreases the loss the most is actually saved and used in the next restart step. This algorithm is summarized in Alg. \ref{alg:restart}.
	
	\textbf{Iterative}: in this version, we have only the internal iteration loop, and the adversarial image is updated each time that an attack decreases the loss value. The algorithm is summarized in Alg. \ref{alg:iterative}.
	
	Intuitively, the first approach requires more iterations, because only the best attacks are saved and preserved, while the second approach updates the adversarial image after each attack. The latter could require less iterations, but it can happen that an attack leads to a region of the search space which is sub-optimal, due to the lack of any controlling strategy.

	\section{Experimental evaluation}
	\subsection{Experimental setup}
	The evaluation of the proposed attack is carried out on the following datasets: CIFAR10 \cite{krizhevsky2009learning}, TinyImagenet, and ImageNet \cite{5206848}. 
	We train CIFAR10 and TinyImagenet using, respectively, ResNet20 \cite{he2016deep} and VGG11 \cite{simonyan2014very} for CIFAR10 and ResNet50 \cite{he2016deep} and VGG16 \cite{simonyan2014very} for TinyImageNet. Regarding ImageNet, we used the pretrained version of ResNet50 \cite{he2016deep} and VGG16 \cite{simonyan2014very}, without fine-tuning them.
	
	As training procedure for CIFAR10 and TinyImageNet, we use Stochastic Gradient Descent, with learning rate set to $0.01$ and momentum to $0.9$. Also, we use the standard augmentation schema: the images are flipped horizontally with a probability of $0.5$, and random cropped to the same original size, but after applying a padding. 
	
	For each experiment, we extract the test images to attack from the subset of test images that are correctly classified by the models. For CIFAR10 we use $100$ images per each class, while for TinyImageNet and ImageNet, we use, respectively, $5$ and $1$ images per class. In this way, we test the attacks on the same number of images for each dataset ($1000$). We also test the ability of the approaches to perform targeted attacks. For this purpose, we use $20$ images per class for CIFAR10, resulting in $200$ attacks, excluding the other datasets.

	We compare our proposal to two others black-box attacks based on the $L_0$ norm: ScratchThat \cite{jere2019scratch} and OnePixel attack \cite{su2019one}. For the latter, we based our implementation on the one present in TorchAttacks \cite{kim2020torchattacks}, while the first one is a custom implementation \footnote{The complete code used to run all the experiments can be found \href{https://github.com/jaryP/PixleAttack}{here}}. For each attack we performed a search of the best hyper-parameters, based on the results from the respective papers, in order to produce the best results in the smallest number of iterations. For ScratchThat, we use $1$ Bézier Curve for CIFAR10 and $3$ otherwise, with a population size of $50$ and maximum number of iterations set to $50$ for the DE algorithm. While for OnePixel, we set the number of pixels to modify equals to $5$, and, regarding the DE algorithm, we set the population size to $100$ solutions and the number of iterations to $50$. Regarding our approach, we use the Restart-Iterative approach as random search with at most $100$ restarts and $50$ iterations per restart cycle. The patches have a size of $3$ pixels, and the mapping function is the random one, which is calculated again at each iteration. All the methods are interrupted when the current image is misclassified by the model, for non targeted attacks, or when the image is classified as the target class, for targeted ones. 
    
    In order to compare the approaches, we use the following metrics:
    \begin{itemize}
        \item Success Rate: it is defined as the percentage of adversarial images that are misclassified by the neural network.
        \item Iterations: calculated image wise, it is the number of times that a model is interrogated while searching the adversarial counterpart of that image. 
        \item $L_0$ norm: the number of pixels that the approach has modified. 
    \end{itemize}
    
	\subsection{Main results}
    	
	\begin{table}[]
    \caption{Results obtained when attacking multiple datasets trained on ResNet and VGG architectures. For each score, we show the mean and the variance, if present, calculated over all the images on that dataset. Best results for each pair dataset-architecture are highlighted in bold.}
    \label{table:results}
    \centering
    \resizebox{1\linewidth}{!}{%
    \begin{tabular}{|c|c|c||c|c|c|}
    \hline 
    Dataset & \multicolumn{1}{c|}{Model} & Method & Success rate & Iterations & $L_0$ norm \\ \hline \hline
    \multirow{6}{*}{\rotatebox[origin=c]{90}{CIFAR10}} & \multirow{3}{*}{\rotatebox[origin=c]{0}{ResNet18}} & OnePixel & \multicolumn{1}{c|}{$64.7$} & \multicolumn{1}{c|}{$5125_{\pm 799}$} & \multicolumn{1}{c|}{$5$} \\ \cline{3-6} 
     & & ScratchThat & $99.7$ & $1010$ & $38.3_{\pm 5.2}$  \\ \cline{3-6}
    & & \textbf{\nome{} (Proposed)} & $\mathbf{100}$ & $119_{\pm 141}$ & $26.8_{\pm 22.8}$  \\  \cline{2-6}
    \noalign{\vskip-1\tabcolsep \vskip-2\arrayrulewidth \vskip\doublerulesep}
    \\ \cline{2-6}
    & \multirow{3}{*}{\rotatebox[origin=c]{0}{VGG11}} & OnePixel & \multicolumn{1}{c|}{$84.5$} & \multicolumn{1}{c|}{$5100$} & \multicolumn{1}{c|}{$5$} \\ \cline{3-6}
     & & ScratchThat & $99.3$ & $1010$ & $27.64_{\pm 4.8}$ \\ \cline{3-6}
     & & \textbf{\nome{} (Proposed)} & $\mathbf{100}$ & $80_{\pm 145}$ & $20.1_{\pm 21.9}$  \\ \hline \hline
    \multirow{6}{*}{\rotatebox[origin=c]{90}{TinyImageNet}} & \multirow{3}{*}{\rotatebox[origin=c]{0}{ResNet50}} & OnePixel & \multicolumn{1}{c|}{$21.0$} & \multicolumn{1}{c|}{$5100$} & \multicolumn{1}{c|}{${5}$} \\ \cline{3-6} 
     & & ScratchThat & $69.7$ & ${1010}$ & $49.1_{\pm 2.2}$  \\ \cline{3-6}
      & & \textbf{\nome{} (Proposed)} & $\mathbf{99.6}$ & $310_{\pm 561}$ & $59.0_{\pm 88.5}$ \\ \cline{2-6}
\noalign{\vskip-1\tabcolsep \vskip-2\arrayrulewidth \vskip\doublerulesep}
\\ \cline{2-6}
    & \multirow{3}{*}{\rotatebox[origin=c]{0}{VGG16}} & OnePixel & \multicolumn{1}{c|}{$31.9$} & \multicolumn{1}{c|}{$5100$} & \multicolumn{1}{c|}{${5}$} \\ \cline{3-6}
     & & ScratchThat & $76.6$ & ${1010}$ & $48.6_{\pm 2.4}$ \\ \cline{3-6}
    & & \textbf{\nome{} (Proposed)} & $\mathbf{100}$ & $87_{\pm 201}$ & $21.5_{\pm 30.6}$ \\ \hline \hline
    \multirow{6}{*}{\rotatebox[origin=c]{90}{ImageNet}} & \multirow{3}{*}{\rotatebox[origin=c]{0}{ResNet50}} & OnePixel & \multicolumn{1}{c|}{$47.7$} & \multicolumn{1}{c|}{$5100$} & \multicolumn{1}{c|}{$5$} \\ \cline{3-6} 
     & & ScratchThat & $82.6$ & $623_{\pm 321}$ & $175.2_{\pm 25.7}$  \\ \cline{3-6}
     & & \textbf{\nome{} (Proposed)} & $\mathbf{98.0}$ & $341_{\pm 426}$ & $155.7_{\pm 184.2}$ \\ \cline{2-6}
\noalign{\vskip-1\tabcolsep \vskip-2\arrayrulewidth \vskip\doublerulesep}
\\ \cline{2-6}
    & \multirow{3}{*}{\rotatebox[origin=c]{0}{VGG16}} & OnePixel & \multicolumn{1}{c|}{$31.9$} & \multicolumn{1}{c|}{$5100$} & \multicolumn{1}{c|}{$5$} \\ \cline{3-6}
     & & ScratchThat & $81.8$ & $753_{\pm 156}$ & $143.0_{\pm 6.0}$ \\ \cline{3-6}
     & & \textbf{\nome{} (Proposed)} & $\mathbf{99.0}$ & $519_{\pm 780}$ & $98.5_{\pm 137.5}$ \\ \hline
    \end{tabular}%
    }
    \end{table}
    
	\subsubsection{Non targeted attacks}
	The results in Table \ref{table:results} shows that our approach is the most performant one across the vast majority of scenarios. In fact, it is capable of achieving almost $100\%$ of success rate on all the datasets, while the others approaches fail to achieve comparable results when the images become bigger.
	
\begin{figure}
     \centering
     \begin{subfigure}[b]{\linewidth}
         \centering
         \includegraphics[width=0.7\textwidth]{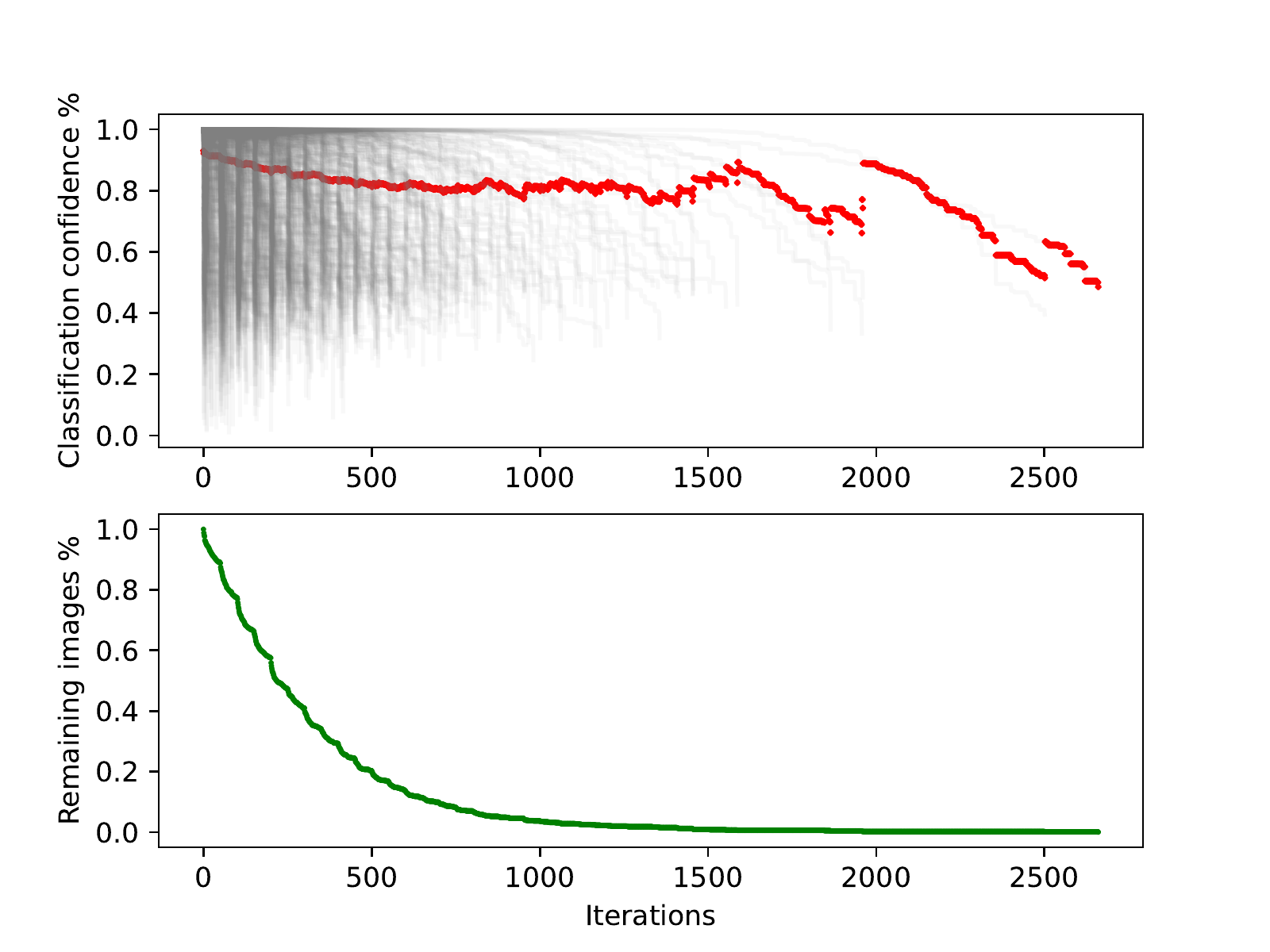}
         \caption{CIFAR10 results using VGG11.}
     \end{subfigure}
    \vfill
     \begin{subfigure}[b]{\linewidth}
         \centering
         \includegraphics[width=0.7\textwidth]{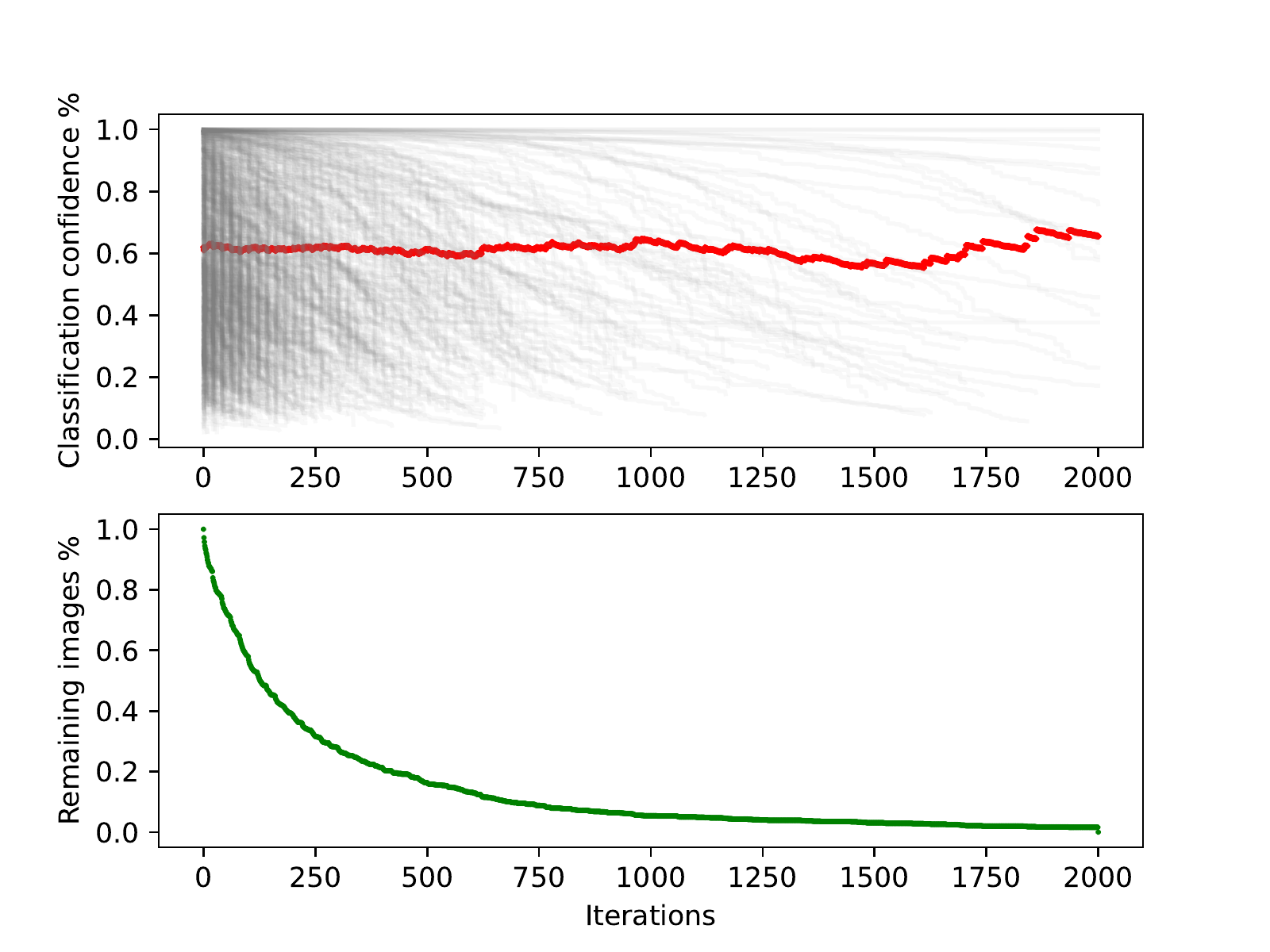}
         \caption{ImageNet results using ResNet50.}
     \end{subfigure}
    \caption{Each figure shows, on top, how the losses (the probability associated to the correct class) change during the iterations of our proposal, using the Restart-Iterative algorithm (the red dots are the average loss calculated on that iteration); while the bottom image shows how many images are left to attack after each iteration. Better viewed in colors.}
    \label{fig:iterations}
    \end{figure}

	Regarding the iterations, we see that our proposal requires a lower number of iterations even when the dimensions of the images increase with the exception of OnePixel, which requires fewer iterations than the others, but the success rate is not sufficiently high to be considered a good attack for big images. Figure \ref{fig:iterations} shows how the loss value decreases when attacking CIFAR10 and ImageNet using our proposal. The images contain the loss value for each attacked image, as well as the average loss on each iteration (red dots). According to the images, it can be seen that a set of images are easily attacked and the associated loss values rapidly decrease, as the bottom images also show; this happens for both CIFAR10 and ImageNet. In fact, for CIFAR10 our approach correctly attacks $80\%$ of the images after $600$ iterations, while the same number of images from ImageNet are correctly attacked after $750$ iterations (approximately).
	
    In the end we analyze the $L_0$ norm. The same table shows that our approach changes less pixels than ScratchThat, when attacking all the dataset. Also, if we cross reference the results in terms of $L_0$ norms and Fig. \ref{fig:iterations}, we see that a lot of images are correctly attacked using a small number of pixels, while other images require more iterations (as also shown by the standard deviation of the metrics). Moreover, the hyperparamenters of our approach can be adapted in order to achieve a smaller $L_0$ loss, at the expense of the number of iterations or success rate (as shown in Table \ref{table:dimensions}).
    
	\subsubsection{Targeted attack}
	
\begin{figure}[h!]
     \centering
     \begin{subfigure}[t]{.45\linewidth}
         \centering
         \includegraphics[width=1\textwidth]{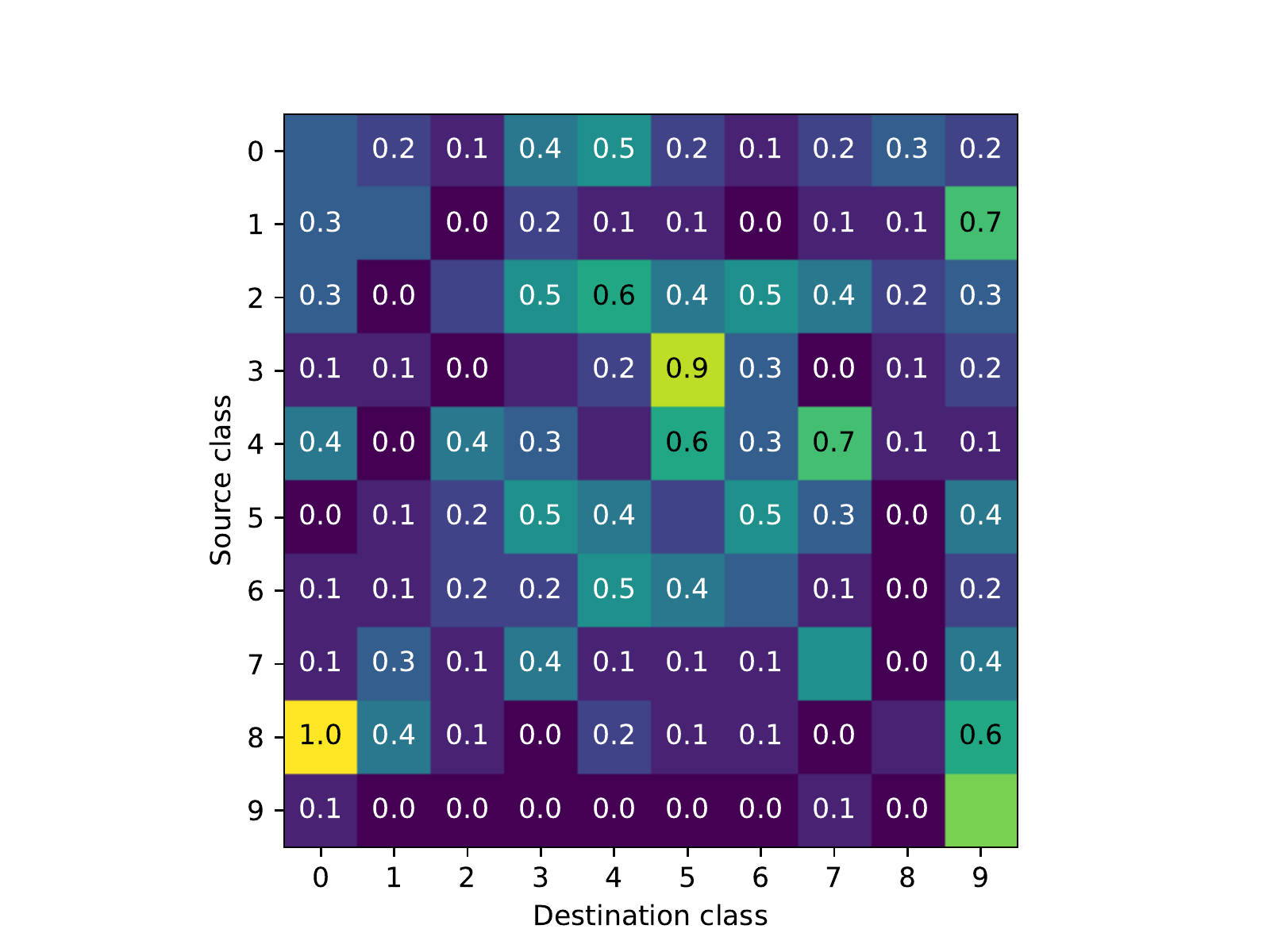}
         \caption{The results obtained using OnePixel attack.}
     \end{subfigure}
     \vfill
     \begin{subfigure}[t]{.45\linewidth}
         \centering
         \includegraphics[width=1\textwidth]{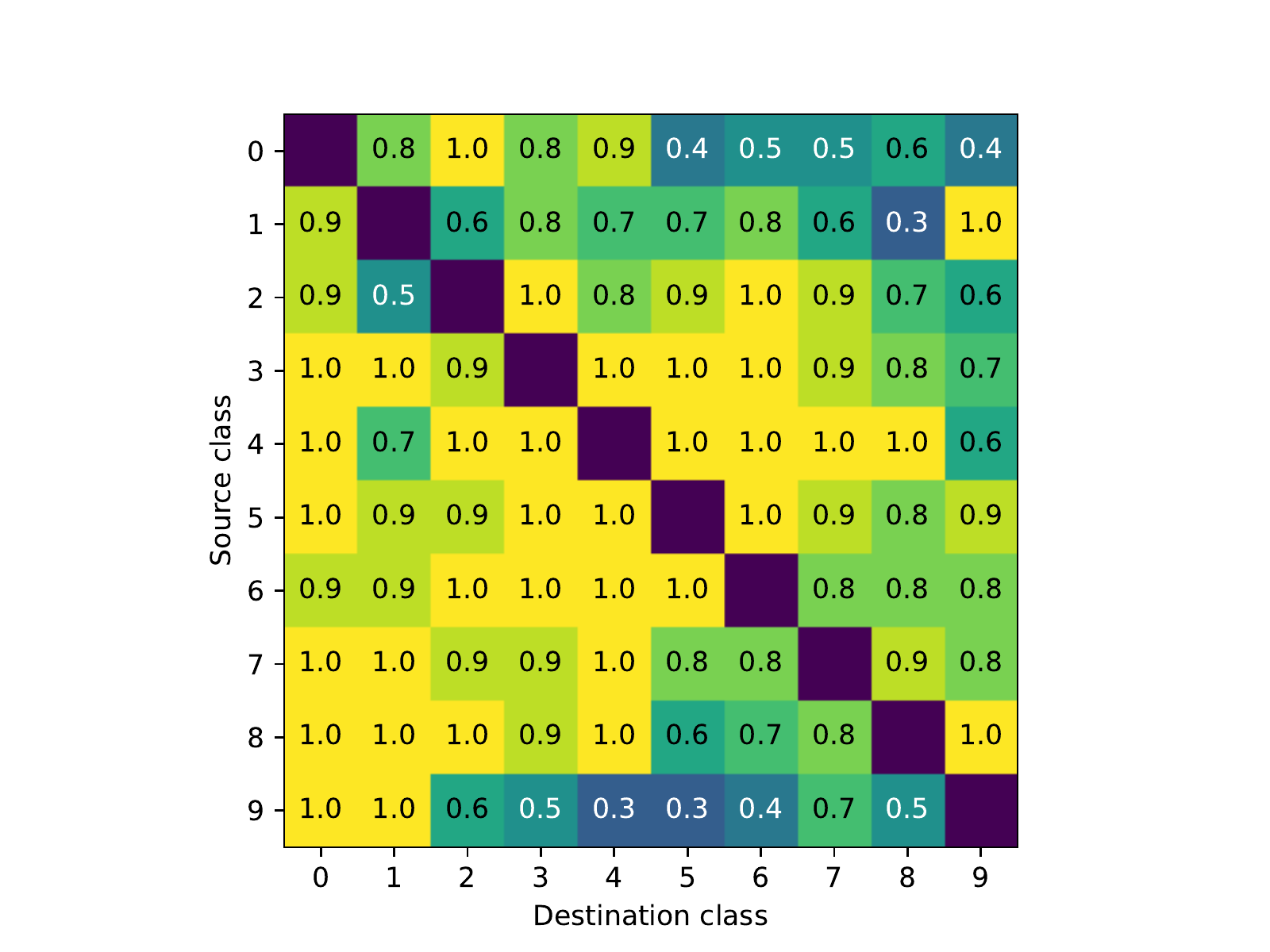}
         \caption{The results obtained using Scratch That attack}
     \end{subfigure}
    \vfill
      \begin{subfigure}[t]{.45\linewidth}
     \centering
     \includegraphics[width=1\textwidth]{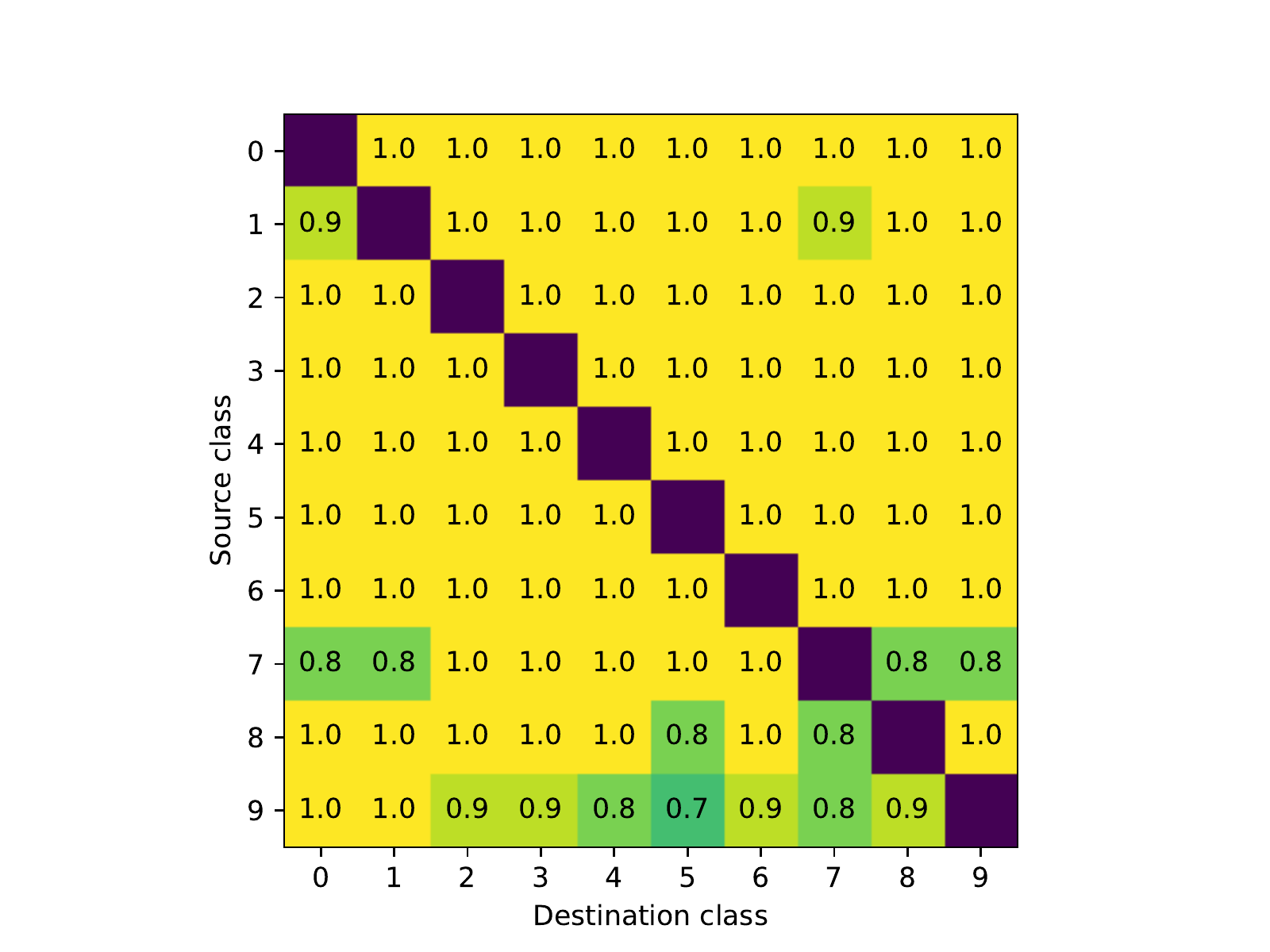}
     \caption{The results obtained using our proposal.}
     \end{subfigure}
    \caption{The images show the success rate on ResNet18 trained using CIFAR10. The results are calculated using $20$ test images, classified correctly by the model, for each class. Each matrix contains the percentage of attacks that have been successfully completed.}
    \label{fig:targeted}
    \end{figure}

	Figure \ref{fig:targeted} shows the results obtained when attacking each class of CIFAR10 with respect to all the others classes. It shows that OnePixel is incapable of attacking most of the classes, and presents a low success rate. Regarding Scratch That, it is capable of perfectly attacking most of the pairs source-destination, achieving half of the times a perfect score. Our approach, instead, achieves a perfect score on almost all the attacks, with the exception of some couples. 
    
    
    \subsubsection{Dimension of the patches}
    
    A crucial aspect of our proposal is the dimension of the patches, and in Table \ref{table:dimensions} we study this aspect, by calculating all the metrics while varying the dimensions of them. For each experiment, we use random mapping and at most $100$ restart, each one composed of $50$ iterations.  
    
    We see that, as expected, the bigger the patches are the fewer iterations are required, but at the same time the distance between the original image and the adversarial counterpart grows. This happens because more pixels are moved at the same time, leading rapidly to a minima, which is accepted due to the callback which interrupts the searching process, that could be reached by moving fewer pixels as well (this is clearly visible by comparing the results obtained using $3$ pixels with the one achieved when using $5$ pixels).

	\begin{table}[]
    \caption{The results obtained when varying the side dimension of a patch, for each combination of dataset and architecture, using \nome{}.}
    \label{table:dimensions}
    \centering
    \resizebox{0.7\linewidth}{!}{%
    \begin{tabular}{|c|c|c||c|c|c|}
    \hline 
    Dataset & \multicolumn{1}{c|}{Model} & Patch dimension & Success rate & Iterations & $L_0$ norm \\ \hline \hline
    \multirow{6}{*}{\rotatebox[origin=c]{90}{CIFAR10}} & \multirow{3}{*}{\rotatebox[origin=c]{0}{ResNet18}} & $1$ & $100$ & $314_{\pm 311}$ & $6.9_{\pm 6.1}$ \\ \cline{3-6} 
     & & $3$ & $100$ & $119_{\pm 141}$ & $26.8_{\pm 22.8}$  \\ \cline{3-6}
    & & $5$ & $100$ & $78_{\pm 123}$ & $51.6_{\pm 44.8}$  \\  \cline{2-6}
    \noalign{\vskip-1\tabcolsep \vskip-2\arrayrulewidth \vskip\doublerulesep}
    \\ \cline{2-6}
    & \multirow{3}{*}{\rotatebox[origin=c]{0}{VGG11}} & $1$ & $99.1$ & $343_{\pm 726}$ & $7.3_{\pm 13.6}$ \\ \cline{3-6}
     & & $3$ & $100$ & $80_{\pm 145}$ & $20.1_{\pm 21.9}$ \\ \cline{3-6}
     & & $5$ & $100$ & $56_{\pm 236}$ & $38.2_{\pm 33.9}$  \\ \hline \hline
    \multirow{6}{*}{\rotatebox[origin=c]{90}{TinyImageNet}} & \multirow{3}{*}{\rotatebox[origin=c]{0}{ResNet50}} & $1$ & $96.81$ & $754_{\pm 1055}$ & $15.3_{\pm 20.0}$ \\ \cline{3-6} 
     & & $3$ & $99.6$ & $310_{\pm 561}$ & $59.0_{\pm 88.5}$  \\ \cline{3-6}
      & & $5$ & $99.8$ & $182_{\pm 379}$ & $ 98.0_{\pm 137.7}$ \\ \cline{2-6}
\noalign{\vskip-1\tabcolsep \vskip-2\arrayrulewidth \vskip\doublerulesep}
\\ \cline{2-6}
    & \multirow{3}{*}{\rotatebox[origin=c]{0}{VGG16}} & $1$ & $99.6$ & $242_{\pm 172}$ & $18.0_{\pm 29.9}$ \\ \cline{3-6}
     & & $3$ & $100$ & $87_{\pm 201}$ & $21.5_{\pm 30.6}$ \\ \cline{3-6}
    & & $5$ & $100$ & $52_{\pm 120}$ & $43.2_{\pm 49.6}$ \\ \hline \hline
    \multirow{6}{*}{\rotatebox[origin=c]{90}{ImageNet}} & \multirow{3}{*}{\rotatebox[origin=c]{0}{ResNet50}} & $1$ & $94.5$ & $1417_{\pm 1376}$ & $28.9_{\pm 27.3}$ \\ \cline{3-6} 
     & & $3$ & $98.0$ & $341_{\pm 426}$ & $155.7_{\pm 184.2}$  \\ \cline{3-6}
     & & $5$ & $99.5$ & $223.3_{\pm 315}$ & $286.2_{\pm 372.3}$ \\ \cline{2-6}
\noalign{\vskip-1\tabcolsep \vskip-2\arrayrulewidth \vskip\doublerulesep}
\\ \cline{2-6}
    & \multirow{3}{*}{\rotatebox[origin=c]{0}{VGG16}} & $1$ & $95.6$ & $1194_{\pm 1319}$ & $24.4_{\pm 26.1}$  \\ \cline{3-6}
     & &$3$ & $99.0$ & $519_{\pm 780}$ & $98.5_{\pm 137.5}$ \\ \cline{3-6}
    &  &$5$ & $99.5$ & $339_{\pm 609}$ & $183.9_{\pm 291.7}$  \\ \hline
    \end{tabular}%
    }
    \end{table}
    
    
    
	\subsection{Ablation studies}
    We perform an ablation study to show how the individual design decisions improve the performance of \nome{}, both in terms of scores and convergence. 
    \label{table:restart_iteration}
	\subsubsection{Restart-Iterative vs Iterative}
    \begin{table*}[t]
    \centering
    \caption{The results obtained using VGG11 trained on CIFAR10. For the Iterative-Restart approach we fixed the number of iterations to $59$.}
    \resizebox{\linewidth}{!}{%
    \begin{tabular}{l|cccc||ccccccc|}
    \cline{2-12}
     & \multicolumn{4}{c||}{Restart-Iterative (restarts)} & \multicolumn{7}{c|}{Iterative (iterations)} \\ \cline{2-12} 
     & \multicolumn{1}{c|}{50} & \multicolumn{1}{c|}{100} & \multicolumn{1}{c|}{250} & 500 & \multicolumn{1}{c|}{50} & \multicolumn{1}{c|}{100} & \multicolumn{1}{c|}{250} & \multicolumn{1}{c|}{500} & \multicolumn{1}{c|}{1000} & \multicolumn{1}{c|}{2000} & 10000 \\  \cline{2-12}
    \noalign{\vskip-1\tabcolsep \vskip-2\arrayrulewidth \vskip\doublerulesep}
    \\ \cline{1-12}
    \multicolumn{1}{|l|}{Success Rate} & \multicolumn{1}{c|}{$95.5$} & \multicolumn{1}{c|}{$98.2$} & \multicolumn{1}{c|}{$99.7$} & $100.0$ & \multicolumn{1}{c|}{$6.1$} & \multicolumn{1}{c|}{$77.7$} & \multicolumn{1}{c|}{$87.9$} & \multicolumn{1}{c|}{$90.5$} & \multicolumn{1}{c|}{$92.4$} & \multicolumn{1}{c|}{$93.1$} & $94.3$ \\ \hline
    \multicolumn{1}{|l|}{Iterations} & \multicolumn{1}{c|}{$170_{\pm{243}}$} & \multicolumn{1}{c|}{$199_{\pm 362}$} & \multicolumn{1}{c|}{$222_{\pm 531}$} & $209_{\pm 468}$ & \multicolumn{1}{c|}{$31_{\pm 17}$} & \multicolumn{1}{c|}{$45_{\pm 36}$} & \multicolumn{1}{c|}{$69_{\pm 79}$} & \multicolumn{1}{c|}{$95_{\pm 147}$} & \multicolumn{1}{c|}{$131_{\pm 264}$} & 
    \multicolumn{1}{c|}{$206_{\pm 511}$} & $697_{\pm 2365}$ \\ \hline
    \multicolumn{1}{|l|}{ $L_0$ norm} & \multicolumn{1}{c|}{$8.9_{\pm 11.6}$} & \multicolumn{1}{c|}{$10.2_{\pm 16.9}$} & \multicolumn{1}{c|}{$10.9_{\pm 22.5}$} & $10.9_{\pm 20.3}$ & \multicolumn{1}{c|}{$12.2_{\pm 9.1}$} & \multicolumn{1}{c|}{$16.1_{\pm 13.7}$} & \multicolumn{1}{c|}{$18.7_{\pm 18.9}$} & \multicolumn{1}{c|}{$20.4_{\pm 23.0}$} & \multicolumn{1}{c|}{$20.4_{\pm 23.4}$} & \multicolumn{1}{c|}{$21.1_{\pm 25.5}$} & $21.86_{\pm 28.2}$ \\ \hline
    \end{tabular}%
    }
    \end{table*}
	
	In this section we explore how the choice of the search algorithm affects the results. Table \ref{table:restart_iteration} shows the results obtained training VGG11 on CIFAR10, while changing the number of iterations. To this end, the algorithm attacks $1$ pixel at a time, using a random mapping calculated at each iteration. 	We can see that the random search based on the restarts is more effective with respect to all the metrics. In fact, when setting the limit to $100$ iterations and $50$ restarts, we achieve better results then the iterative algorithm with a limit of $10000$ iterations. 
	

    \subsubsection{Mapping function}
    
	\begin{table}[]
    \caption{The results obtained when varying the mapping function. The attacked model is VGG11 trained on CIFAR10.}
    \label{table:mapping}
    \centering
    \resizebox{0.5\linewidth}{!}{%
    \begin{tabular}{|c||c|c|c|}
    \hline 
    Mapping function & Success rate & Iterations & $L_0$ norm \\ \hline \hline
     Distance & $17.6$ & $8297_{\pm 3697}$ & $40.1_{\pm 17.9}$ \\ \hline
     Similarity & $96.5$ & $1029_{\pm 2196}$ & $10.0_{\pm 18.7}$  \\ \hline
     Distance distribution & $99.3$ & $536_{\pm 1260}$ & $5.9_{\pm 11.9}$ \\ \hline
     Similarity distribution & $99.1$ & $605_{\pm 1425}$ & $3.1_{\pm 7.3}$ \\ \hline
    \end{tabular}%
    }
    \end{table}
    
    In this section we study how the selection of mapping function affects the final scores, shown in Table \ref{table:mapping}. We see that, when operating with a deterministic mapping, the number of required iterations, as well as the $L_0$ norm, are higher than the distribution counterparts. This happens because the randomness of the distributions force the approach to explore the space, avoiding local minima.
    
    \section{Conclusion}
    
    In this paper we proposed a novel black-box attack called \nome{}. Our attack, despite using only the probability of the prediction returned by the attacked model, proved to be effective over a wide range of datasets and architectures types, and on multiple metrics, such as the number of iterations required and the number of modified pixels. 
    
    As future work, we aim to understand the correlation between the attacked class and the pixels in the image, especially when performing a targeted attack. Furthermore, we would like to expand the proposed approach also on other domains, such as text and audio. In the end, we would like to understand if the proposed attack can be further studied, in order to create a defense algorithm against $L_0$ attacks.  

    \bibliographystyle{elsarticle-num}
	\bibliography{main}
    
\end{document}